\documentclass[10pt,twocolumn,letterpaper]{article}

\usepackage{cvpr}      
\usepackage{wrapfig}

\usepackage[table]{xcolor}
\usepackage{multirow}
\usepackage{amsmath}            
\usepackage{algorithm}
\usepackage[noend]{algpseudocode}

\definecolor{cvprblue}{rgb}{0.21,0.49,0.74}
\usepackage[pagebackref,breaklinks,colorlinks,allcolors=cvprblue]{hyperref}

\def\paperID{****} 
\def\confName{CVPR}
\def\confYear{2026}

\title{SVC 2026: the Second Multimodal Deception Detection Challenge and the First Domain Generalized Remote Physiological Measurement Challenge}

\author{
\centering
\begin{tabular}{c}
Dongliang Zhu$^{1}$\thanks{Equal contribution.} \quad
Zhiyi Niu$^{6}$\footnotemark[1] \quad
Bo Zhao$^{2}$\footnotemark[1] \quad
Jiajian Huang$^{2}$\footnotemark[1] \quad
Shuo Ye$^{3}$\footnotemark[1] \quad
Xun Lin$^{4}$\footnotemark[1] \\[0.4ex]
Hui Ma$^{2}$\footnotemark[1] \quad
Taorui Wang$^{2}$ \quad
Jiayu Zhang$^{2}$ \quad
Chunmei Zhu$^{5}$ \quad
Junzhe Cao$^{2}$ \quad
Yingjie Ma$^{2}$ \\[0.4ex]
Rencheng Song$^{6}$ \quad
Albert Clap\'es$^{7}$ \quad
Sergio Escalera$^{7}$ \quad
Dan Guo$^{6}$ \quad
Zitong Yu$^{2}$\thanks{Corresponding author.} \\[1.0ex]
$^{1}$ Wuhan University \quad
$^{2}$ Great Bay University \quad
$^{3}$ Tsinghua University \\[0.3ex]
$^{4}$ The Chinese University of Hong Kong \quad
$^{5}$ Sun Yat-sen University \\[0.3ex]
$^{6}$ Hefei University of Technology \quad
$^{7}$ University of Barcelona
\end{tabular}
}

\begin{document}
\maketitle
\begin{abstract}
Subtle visual signals, although difficult to perceive with the naked eye, contain important information that can reveal hidden patterns in visual data. These signals play a key role in many applications, including biometric security, multimedia forensics, medical diagnosis, industrial inspection, and affective computing. With the rapid development of computer vision and representation learning techniques, detecting and interpreting such subtle signals has become an emerging research direction. However, existing studies often focus on specific tasks or modalities, and models still face challenges in robustness, representation ability, and generalization when handling subtle and weak signals in real-world environments. To promote research in this area, we organize the Subtle visual Challenge, which aims to learn robust representations for subtle visual signals. The challenge includes two tasks: cross-domain multimodal deception detection and remote photoplethysmography (rPPG) estimation. We hope that this challenge will encourage the development of more robust and generalizable models for subtle visual understanding, and further advance research in computer vision and multimodal learning. A total of 22 teams submitted their final results to this workshop competition, and the corresponding baseline models have been released on the \href{https://sites.google.com/view/svc-cvpr26}{MMDD2026 platform}\footnote{\url{https://sites.google.com/view/svc-cvpr26}}

\end{abstract}    
\vspace{-1.3em}
\section{Introduction}
\label{sec:intro}

Subtle visual signals are often imperceptible to the human eye, yet they contain critical information that reveals underlying patterns in visual data. With the support of computer vision and representation learning techniques, these signals can be effectively modeled to improve the understanding of complex environments. Such signals are typically characterized by extremely low amplitude, short duration, and high sensitivity to noise. Representative examples include slight facial color variations, subtle muscle movements, and fine-grained temporal dynamics. Although difficult to perceive directly, these signals encode important cues related to an individual’s physiological and psychological states, making them valuable for applications such as biometric security, multimedia forensics, healthcare monitoring, and affective computing. However, in real-world scenarios, these weak signals are easily affected by illumination changes~\cite{2023cross}, motion artifacts, and device variations, posing significant challenges for stable modeling and robust representation learning.

\begin{figure}[t]
	\centering
	\includegraphics[width=0.45\textwidth]{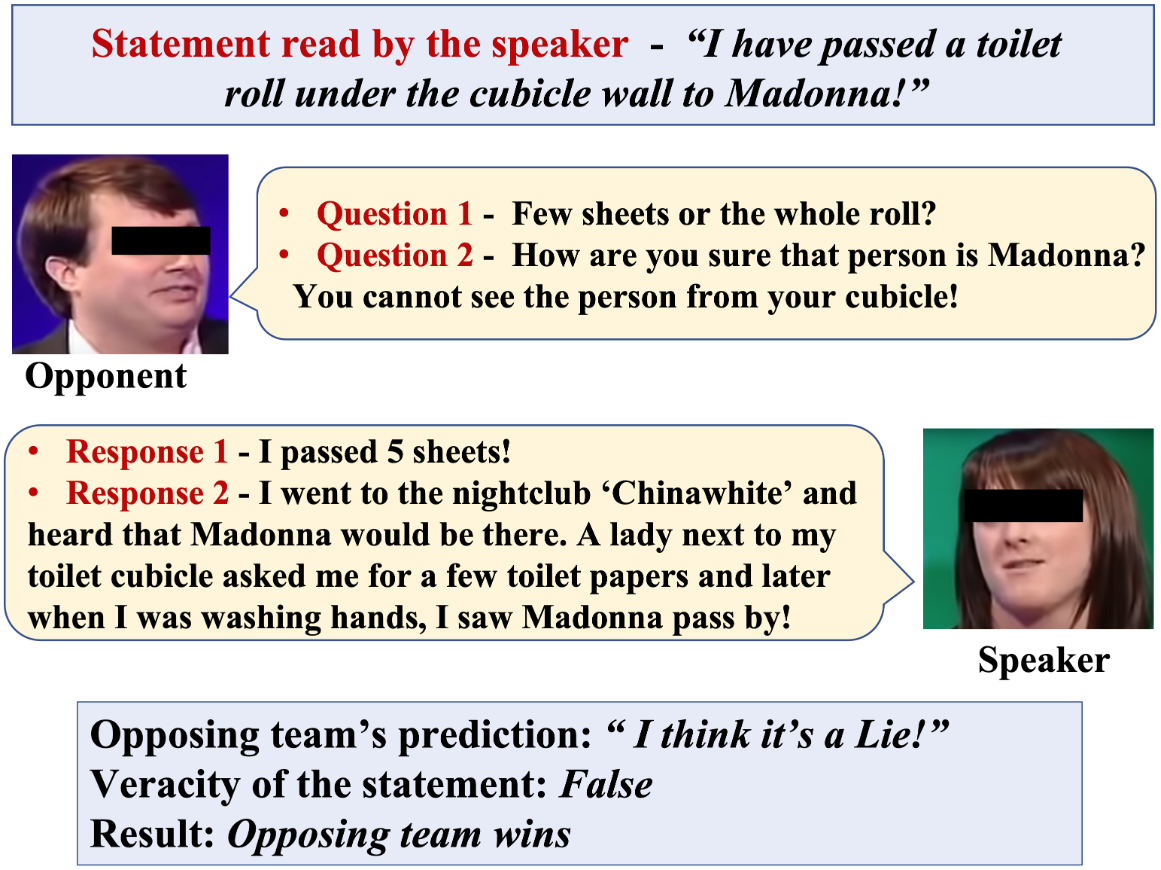}
	\caption{Examples of deceptive actions.}
	\label{dolos}
    \vspace{-0.2cm}
\end{figure}

In practical applications, deception detection and remote photoplethysmography (rPPG) are two representative tasks that rely on modeling subtle visual signals. Deception detection aims to identify hidden deceptive cues from multimodal behavioral signals, including facial expressions, body movements, and speech. An example of deceptive behavior is illustrated in Fig.~\ref{dolos}. However, deceptive behaviors are inherently concealed, and individuals often deliberately regulate their observable actions to reduce the risk of exposure. Psychological studies suggest that humans tend to exhibit a default bias toward believing others, with deception detection accuracy only slightly above chance level. This limitation makes manual observation insufficient for real-world deployment and motivates the development of automated approaches based on computer vision, multimodal learning, and deep neural networks. Although recent deep learning methods have achieved notable progress, their performance largely depends on the data distribution. When models are transferred to new datasets or real-world scenarios, significant domain shifts often lead to performance degradation. Differences in acquisition conditions, behavioral expressions, and modality distributions across datasets make cross-domain generalization a critical bottleneck for practical applications.

On the other hand, rPPG aims to recover physiological signals from videos by capturing extremely subtle color changes caused by blood circulation. This task is highly sensitive to illumination conditions, head motion, and device variations, which imposes strict requirements on model robustness and generalization. Although rPPG and deception detection differ in their objectives, both fundamentally rely on accurately modeling weak visual signals and face similar challenges in cross-domain and real-world settings. Therefore, studying representation learning for subtle visual signals from a unified perspective can benefit both tasks. Integrating deception detection and rPPG within a common framework also enables a more systematic evaluation of a model’s ability to capture and interpret subtle signals.

Motivated by the above observations, we extend the first Multimodal Deception Detection Challenge (SVC 2025) and introduce the second Visual Subtlety Challenge (SVC 2026). This challenge encourages researchers to develop robust models for subtle visual signal modeling under complex environments and distribution shifts, while leveraging multimodal information to improve deception recognition. Participants are required to submit trained models, checkpoints, well-documented source code, and technical reports describing their methods and results. All submissions will be independently reproduced and verified by the organizers. The final ranking is determined based on multiple evaluation metrics on the test set. In summary, the main contributions of this challenge are as follows:
\begin{itemize}
	\item We establish a unified evaluation framework that integrates deception detection and rPPG, enabling a systematic assessment of a model’s ability to capture subtle visual signals and maintain robustness under real-world.

    \item Building upon the SVC 2025, we extend it to the SVC 2026, with a focus on cross-domain generalization and real-world applicability.

    \item We introduce the first Domain Generalized Remote Physiological Measurement Challenge (PhysDG), with a focus on evaluating robust, domain-invariant rPPG algorithms under strict cross-domain generalization.

    \item We introduce a standardized evaluation protocol, including multiple metrics (accuracy, F1 score, and error rate) and an independent reproducibility pipeline, ensuring fair comparison and reliable validation of submitted methods.

\end{itemize}

\section{Related work}
\label{sec:related}

\subsection{Multimodal Deception Detection.}
In recent years, deception detection has made significant progress. An increasing number of studies integrate verbal and non-verbal cues through feature fusion techniques~\cite{2024fusionmamba}, leveraging multimodal data to improve detection performance~\cite{2026survey}. For example, Wu et al. exploit diverse modalities, including micro-expressions and IDT from RGB videos,  MFCC from audio, and textual transcripts~\cite{57}. Building upon this line of research, Guo et al. propose a parameter-efficient cross-modal learning framework, termed PECL, which combines a Unified Temporal Adapter (UT-Adapter) with a plug-and-play audiovisual fusion module (PAVF), along with a multi-task learning strategy, achieving both efficient and accurate deception detection~\cite{10}. Furthermore, Zhu et al., inspired by micro-gesture analysis~\cite{Micro-Gesture}, introduce a dynamic learning framework, namely DLF-BRAM. This approach segments head and limb regions from videos and performs relation-aware modeling to capture coordinated patterns across different body parts~\cite{2025detecting}.

Despite these advances, cross-domain generalization remains a key challenge in multimodal deception detection. Existing works mainly focus on modeling and optimization within a single domain, while paying limited attention to the transferability of models across different scenarios or populations. For instance, models trained on controlled laboratory datasets often fail to generalize to real-world environments, where variations in recording conditions, behavioral expressions, and interaction patterns introduce significant domain shifts.

To address this issue, this challenge follows the protocol of SVC 2025~\cite{2025svc} and evaluates cross-domain generalization on widely used audiovisual deception detection datasets. This setting enables a systematic assessment of model robustness across diverse scenarios, facilitates fair comparison among methods, and highlights current limitations, thereby promoting the development of cross-domain capable deception detection systems.

\subsection{Remote Rhotoplethysmography}
Remote photoplethysmography (rPPG) is a promising non-contact technology that can estimate vital signs such as heart rate directly from facial videos. Existing research in this field mainly focuses on two directions: optimizing feature learning paradigms and improving domain adaptation performance. In terms of feature learning, early methods relied on handcrafted algorithms to capture subtle skin color variations caused by pulse. With the development of deep learning technology, the research focus has shifted to constructing sophisticated spatiotemporal networks and physics-guided architectures to extract robust deep rPPG representations ~\cite{qian2024dual, li2023channel, zhao2025phase}. Subsequent technological innovations have further enriched the learning methods of feature representations, including separating physiological dynamic features from structural noise ~\cite{qian2025physdiff}, mining inherent physiological clues in human faces ~\cite{qian2024cluster}, and integrating multimodal sensor data or large language models to expand the contextual dimension of physiological sensing ~\cite{wu2025cardiacmamba, xie2025physllm}. In addition, privacy-preserving data editing techniques have been introduced into this field to artificially expand the distribution range of training data ~\cite{tu2026phys}.

Although relevant research methods have achieved considerable progress, extracting reliable physiological signals in various real-world scenarios remains a key challenge in this field. Affected by domain shifts caused by uncontrolled illumination, diverse camera sensors, different skin tones and motion artifacts, the performance of most methods degrades significantly. To alleviate this problem, another line of research captures domain-invariant physiological features through domain adaptation and meta-learning strategies. However, these strategies often fail to maintain stable and accurate performance when generalizing to completely unseen domains, especially under the common constraint of fixed model weights in real-world deployment scenarios. In response to the above problems, the Domain Generalized Remote Physiological Measurement (PhysDG) Challenge has constructed a comprehensive two-phase evaluation framework, including multi-domain fusion training and fixed model weights. By assessing the model's generalization ability to unseen samples in the same domain and cross-domain generalization ability under stringent conditions, the challenge aims to make up for the deficiencies in the current technological deployment and promote the development of domain-invariant rPPG algorithms with good robustness.
\section{The Second Multimodal Deception Detection Challenge}

\subsection{Challenge Corpora}
This challenge adopts four publicly available datasets for training: Real-life Deception Detection (Real-life Trial), Bag-of-Lies, Box-of-Lies, and the Miami University Deception Detection Database (MU3D). Detailed statistics are summarized in Table~\ref{tab:datasets}. All participants are required to sign data usage agreements and obtain access to the original datasets from their respective official platforms. The organizers do not directly distribute raw data; instead, we provide pre-extracted features, including OpenFace-based facial features, emotion representations extracted from pre-trained models, and Mel-spectrograms generated using PyTorch. These features are anonymized and do not contain any personally identifiable information.

\begin{figure}[ht]
	\centering
	\includegraphics[width=0.48\textwidth]{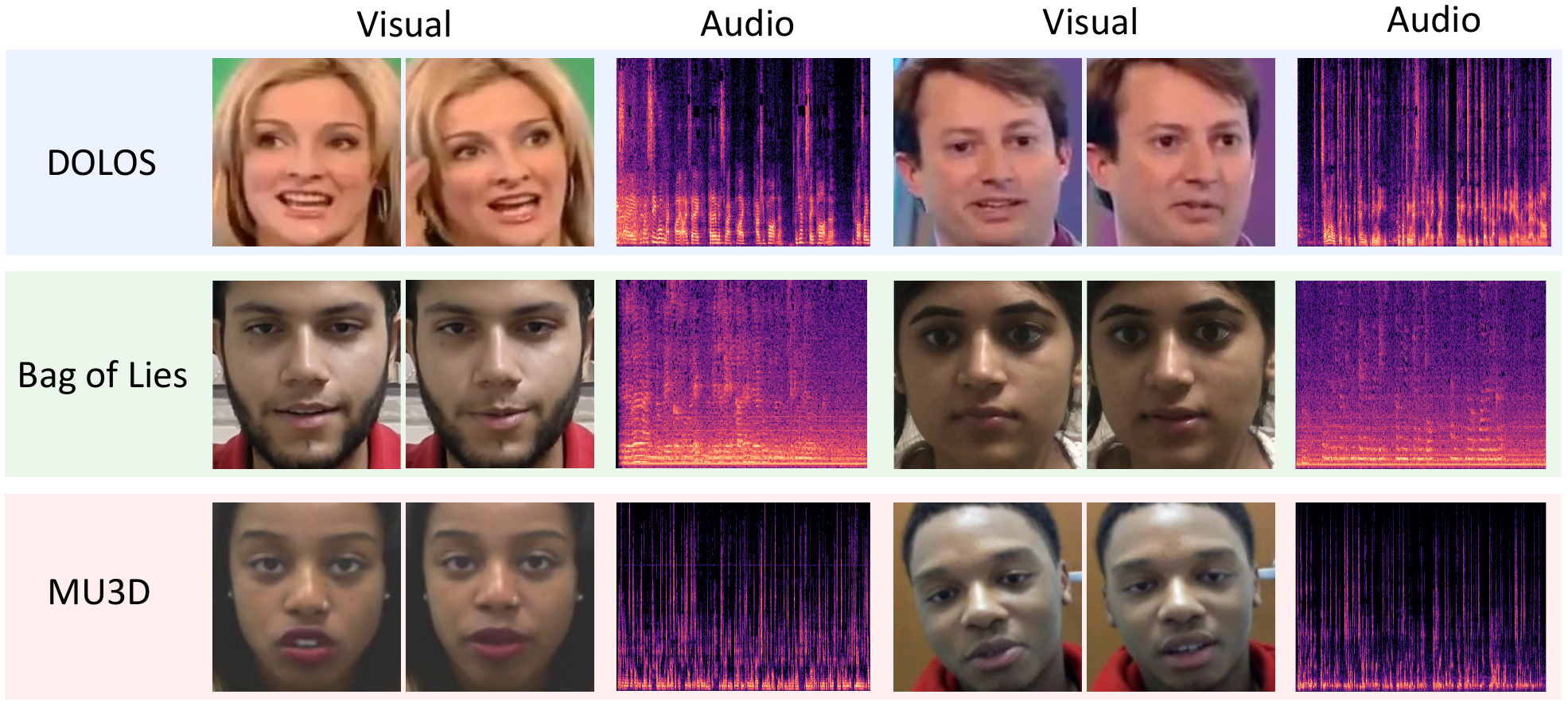}
	\caption{Sample examples from the DOLOS, Bag-of-Lies, and MU3D datasets.}
	\label{data_exp}
\end{figure}

\begin{table*}[t] 
\centering
\small
\caption{Multimodal deception detection datasets used in the challenge.}
\begin{tabular}{lcccccc}
\toprule
\textbf{Dataset} & \textbf{\#Subjects} & \textbf{Total} & \textbf{Deceptive} & \textbf{Truthful} & \textbf{Dec/Truth} & \textbf{Scenario} \\
\midrule
Real Life Trials~\cite{3} & 56 & 121 & 61 & 60 & 1.02 & Courtroom \\
Bag of Lies~\cite{11} & 35 & 325 & 162 & 163 & 0.99 & Lab \\
MU3D~\cite{18} & 80 & 320 & 160 & 160 & 1.00 & Lab \\
Box of Lies~\cite{65} & 26 & 1049 & 862 & 187 & 4.61 & Gameshow \\
DOLOS~\cite{10} & 213 & 1675 & 899 & 776 & 1.16 & Gameshow \\
\bottomrule
\end{tabular}

\label{tab:datasets}
\end{table*}

The Real-life Trials dataset~\cite{3} consists of 121 video clips from real courtroom trials, with an average duration of 28 seconds. The dataset includes high-profile cases such as the Jodi Arias trial, testimonies from the Innocence Project, and defendant statements from crime-related television programs. All clips correspond to statements made by defendants or witnesses. Based on judicial outcomes (guilty verdict, not guilty verdict, or exoneration), 61 clips are labeled as deceptive and 60 as truthful. The dataset involves 56 speakers (21 female and 35 male), aged between 16 and 60. It also provides manually verified crowd-sourced transcripts (8,055 words, including fillers and repetitions) and annotations of nine categories of nonverbal behaviors (e.g., facial expressions and hand gestures) based on the MUMIN coding scheme.

The Bag-of-Lies dataset~\cite{11} contains 325 annotated recordings from 35 participants, including 162 deceptive and 163 truthful samples. It integrates four modalities: (1) facial and body expression videos recorded via smartphone, (2) speech audio, (3) gaze tracking data collected using the Gazepoint GP3 eye tracker (including fixation and pupil features), and (4) EEG signals recorded using a 13-channel Emotiv EPOC+ headset at 128 Hz. During data collection, participants were asked to freely describe 6–10 images and independently choose to tell the truth or lie for each image. The duration of recordings ranges from 3.5 to 42 seconds.

The Miami University Deception Detection Database (MU3D)~\cite{18} includes 320 video clips from 80 participants of diverse ethnic backgrounds. Each participant records 4 videos under controlled laboratory conditions: positive truth (describing a person they genuinely like), negative truth (describing a person they genuinely dislike), positive lie (describing a disliked person as liked), and negative lie (describing a liked person as disliked). Each recording lasts approximately 45 seconds.

The Box-of-Lies dataset~\cite{65} is constructed from 25 publicly available episodes of The Tonight Show Starring Jimmy Fallon, totaling 2 hours and 24 minutes of video. It contains 1,049 annotated utterances involving interactions between the host and 26 guests (6 male and 20 female). The dataset captures both truthful and deceptive behaviors in a natural conversational setting. The data processing pipeline includes: extracting videos from YouTube, segmenting utterances using ELAN, annotating multimodal behaviors (e.g., facial expressions, head movements, and gaze) based on the MUMIN scheme, and generating transcripts via Amazon Mechanical Turk followed by manual verification. Each utterance is labeled with a truthfulness tag, including 862 deceptive and 187 truthful samples. Linguistic features are extracted from transcripts, while nonverbal behaviors (e.g., smiling frequency) are quantified as temporal proportions.

For evaluation, we adopt the test split of the DOLOS dataset~\cite{10} as the Stage-1 benchmark. This dataset contains 1,675 video clips from 213 participants (141 male and 72 female), making it one of the largest deception detection datasets in unconstrained, real-world settings. The clip durations range from 2 to 19 seconds, with most samples concentrated in short (2–4 seconds) and medium (5–10 seconds) intervals. It includes 899 deceptive and 776 truthful samples, resulting in a relatively balanced ratio of 1.16 (compared to 4.61 in Box-of-Lies), which reduces sample bias. The data are collected from a UK-based reality comedy game show on YouTube, under fair use policies. All subjects are public figures, and the dataset has received ethical approval (IRB-2022-901). In this challenge, we follow Protocol-I and use its test split, which contains 382 samples, as the Stage-1 evaluation set.
\subsection{Evaluation Metrics}

In this challenge, three core evaluation metrics are adopted: Accuracy, Error Rate, and F1 score. Among them, Accuracy serves as the primary ranking criterion. All metrics are computed based on binary classification results, with label definitions as follows:
\begin{itemize}
    \item Truthful samples are labeled as 1
    \item Deceptive samples are labeled as 0
\end{itemize}

\paragraph{(1) Accuracy}
Accuracy measures the proportion of correctly classified samples over the total number of samples:
\begin{equation}
\text{Accuracy} = \frac{TP + TN}{TP + TN + FP + FN}
\end{equation}
where:
\begin{itemize}
    \item $TP$ (True Positive): correctly predicted deceptive samples
    \item $TN$ (True Negative): correctly predicted truthful samples
    \item $FP$ (False Positive): truthful samples incorrectly predicted as deceptive
    \item $FN$ (False Negative): deceptive samples incorrectly predicted as truthful
\end{itemize}

\paragraph{(2) Error Rate}
Error Rate is the complement of Accuracy, representing the proportion of misclassified samples:
\begin{equation}
\text{Error Rate} = 1 - \text{Accuracy} = \frac{FP + FN}{TP + TN + FP + FN}
\end{equation}

\paragraph{(3) F1 Score}
The F1 score is the harmonic mean of Precision and Recall:
\begin{equation}
\text{Precision} = \frac{TP}{TP + FP}, \quad
\text{Recall} = \frac{TP}{TP + FN}
\end{equation}
\begin{equation}
\text{F1} = \frac{2 \cdot \text{Precision} \cdot \text{Recall}}{\text{Precision} + \text{Recall}}
\end{equation}

\paragraph{Prediction Format}
The model outputs a continuous score in the range $[0,1]$, representing the probability that a sample is deceptive. During evaluation, a threshold of 0.5 is applied to obtain binary predictions:
\begin{equation}
\hat{y} =
\begin{cases}
0, & \text{if } s \geq 0.5 \ (\text{deceptive}) \\
1, & \text{if } s < 0.5 \ (\text{truthful})
\end{cases}
\end{equation}
where $s$ denotes the predicted score.

Participants are required to submit a prediction file containing the sample identifier and the corresponding predicted score. An example format is shown below:
\begin{verbatim}
SJ_BOL_EP3_lie_4 0.14431
\end{verbatim}
\subsection{ Baseline model}

\begin{figure}[t]
	\centering
	\includegraphics[width=0.48\textwidth]{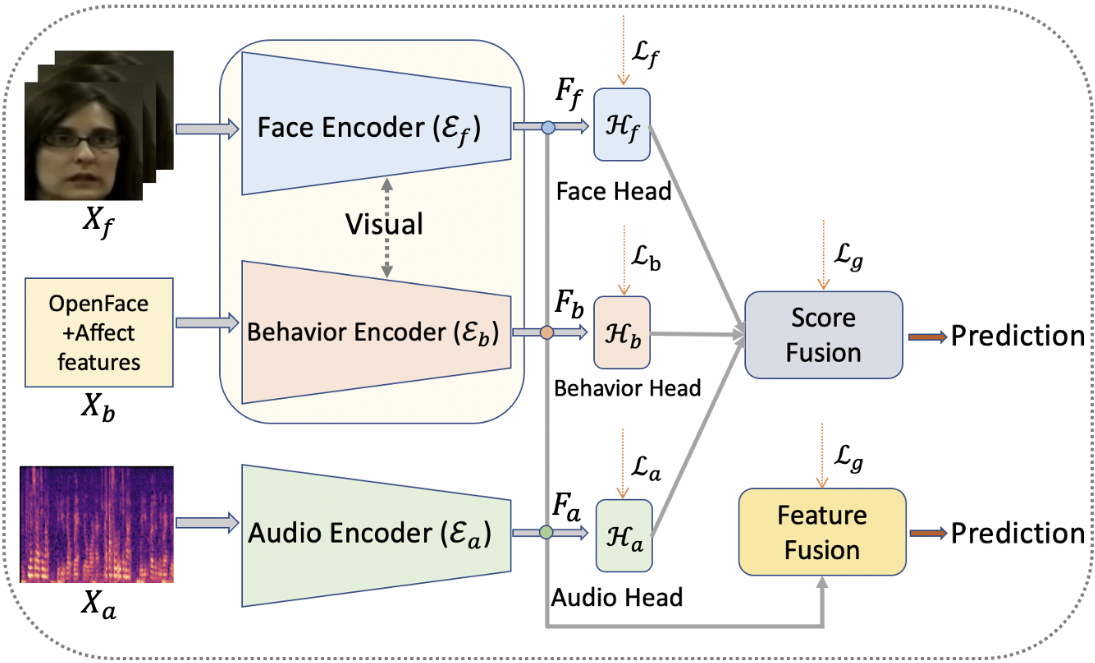}
	\caption{Baseline model.}
	\label{Baseline model}
\end{figure}

We provide a baseline model for cross-domain audiovisual deception detection. The model first extracts frame-level facial features using ResNet18. Behavioral cues are further obtained via OpenFace and EmotionNet, including facial action units, gaze information, and emotion representations. For the audio modality, Mel-spectrogram features are extracted using OpenSmile, while raw waveforms can alternatively be encoded using Wave2Vec. The multimodal features are then projected through linear layers and encoded using Transformer-based modules for unified representation learning, followed by a classifier for final prediction. The framework of the proposed method is shown in Fig. \ref{Baseline model}.

To improve cross-domain generalization, we propose a \textbf{Multimodal Inter-domain Gradient Matching (MM-IDGM)} algorithm. The key idea is to maximize the inner product between gradients of different modality encoders, enforcing alignment of gradient directions across domains and promoting consistent multimodal optimization. Specifically, the method builds upon the Fish algorithm by introducing a dynamic learning rate adjustment strategy. In addition, historical unimodal losses are incorporated to guide the optimization process, further improving generalization performance in the Multi-to-Single setting.

Furthermore, we design an attention-based hybrid fusion module that integrates MLP-Mixer and self-attention mechanisms. The unimodal MLP layers are used to model dynamic variations within each modality, while self-attention captures intra-modal dependencies. Cross-modal MLP layers are introduced to model interactions between different modalities. The fusion module consists of six stacked hybrid attention layers, combined with layer normalization and multi-head self-attention. Input features are progressively projected and transformed, followed by global average pooling to obtain the final representation. This design enhances both intra-domain and inter-domain feature interactions, leading to improved multimodal fusion and overall robustness.
\subsection{ Participation}
A total of 18 teams submitted their results by the end of the workshop competition \footnote{MMDD 2026 Challenge Results: \url{https://www.codabench.org/competitions/12678/\#/results-tab}}. The results of phase 2 are shown in Table. 2. We will introduce the approaches of some teams in the following part.
\begin{table}[t]
\centering
\caption{MMDD Challenge Results.}
\begin{tabular*}{\linewidth}{@{\extracolsep{\fill}} c l c c c}
\toprule
\# & Team & ACC & F1 & ERR \\
\midrule
1 & xkxkxk & 71.35 & 63.9  & 28.65 \\
2 & sqd    & 57.62 & 7.69  & 42.38 \\
3 & ahrior & 57.22 & 11.31 & 42.78 \\
\bottomrule
\end{tabular*}
\label{tab:challenge_results}
\end{table}
\subsubsection{ Team xkxkxk }
\begin{figure}[ht]
	\centering
	\includegraphics[width=0.48\textwidth]{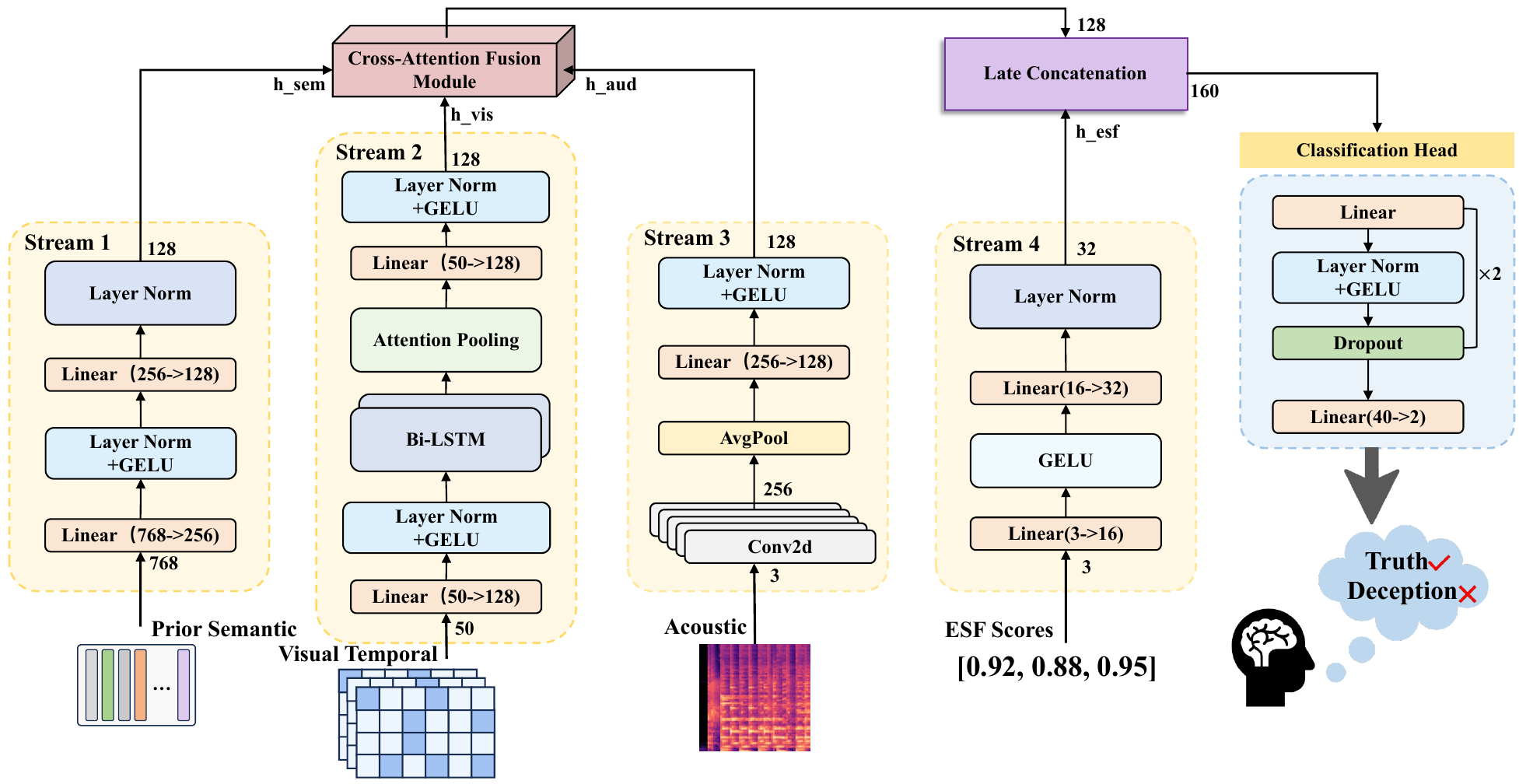}
	\caption{Network framework of Team xkxkxk.}
	\label{xkxkxk}
\end{figure}

We propose a four-stream multimodal fusion framework for audio-visual deception detection, built around a key insight: facial action unit (AU) features from OpenFace and emotion probability vectors from Affect carry explicit semantic meaning — each dimension corresponds to a specific, interpretable behavioral signal (e.g., AU12 = Lip Corner Puller intensity, AU45 = Blink frequency, Affect dimensions = discrete emotion probabilities). In low-data regimes, training neural networks to implicitly re-discover these known semantics from raw numerical features is both data-inefficient and prone to overfitting.
Instead, we introduce a rule-based feature-to-language bridge: temporal AU statistics and emotion distributions are aggregated per time window and translated into structured natural language descriptions via calibrated semantic mappings (e.g., z-score thresholds mapped to intensity labels such as "clearly active" or "elevated"). These formatted prompts are then analyzed by a large language model (LLM), which leverages its pre-trained understanding of facial behavior and psychological cues to produce (1) a rich natural language analysis paragraph capturing high-level behavioral patterns related to deception, and (2) a 3-dimensional Emotional State Feature (ESF) vector quantifying cognitive load, emotion conflict, and suppression. The framework of the proposed method is shown in Fig. \ref{xkxkxk}.

\subsubsection{ Team sqd}
To address the significant domain shifts in multimodal deception detection, Team sqd proposed a novel frequency-aware feature adaptation framework. The architecture processes visual information through two distinct branches: a behavior branch utilizing OpenFace Affect to extract structured facial cues from video inputs, and a spatiotemporal branch employing a ResNet18-GRU backbone to encode raw sequential facial inputs.

\begin{figure}[ht]
	\centering
	\includegraphics[width=0.48\textwidth]{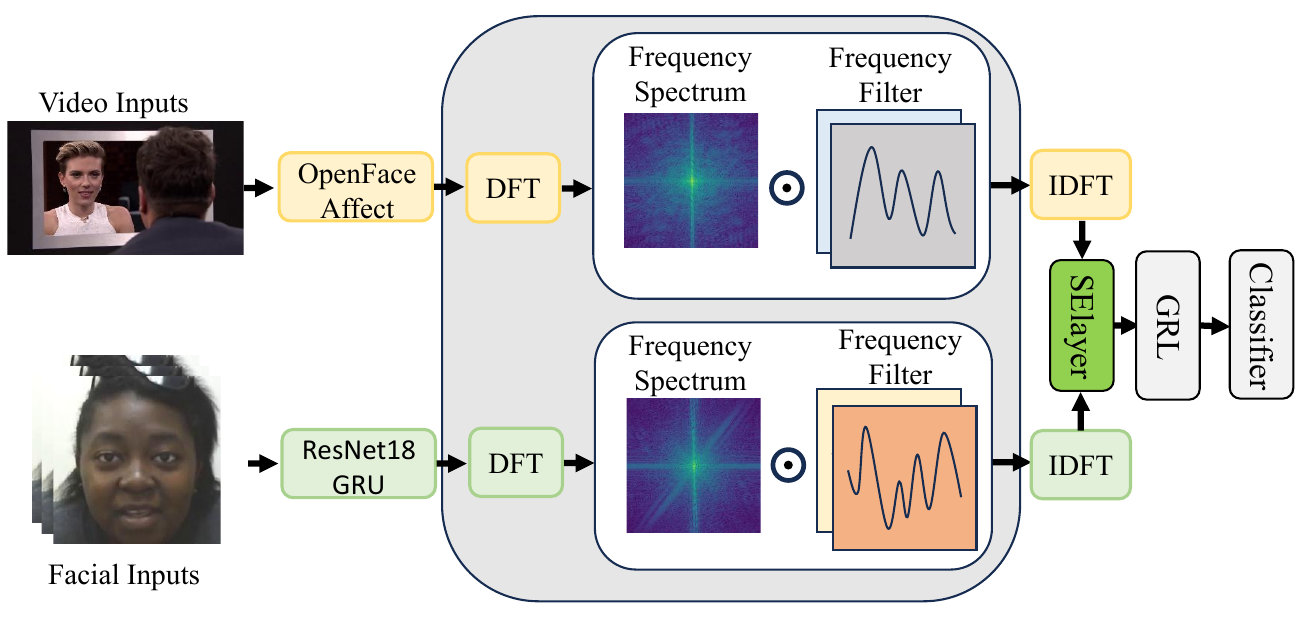}
	\caption{Network framework of Team sqd.}
	\label{sqd}
\end{figure}

To effectively isolate domain-invariant deceptive signals from domain-specific noise, the extracted features from both branches are transformed into the frequency domain via Discrete Fourier Transform (DFT). Customized frequency filters are then applied to the resulting spectra to suppress redundant background information and emphasize generalizable behavioral cues. The filtered signals are subsequently reconstructed back to the spatial-temporal domain using Inverse Discrete Fourier Transform (IDFT). The refined features from both streams are integrated and processed through a Squeeze-and-Excitation (SE) layer to dynamically recalibrate channel-wise feature importance. Finally, inspired by adversarial learning principles, a Gradient Reversal Layer (GRL) is incorporated immediately prior to the multi-class classifier. This facilitates adversarial domain adaptation, ensuring that the final learned representations are both discriminative for veracity and invariant across heterogeneous datasets. The framework of the proposed method is shown in Fig. \ref{sqd}.

\subsubsection{ Team ahrior}

\begin{figure}[ht]
	\centering
	\includegraphics[width=0.48\textwidth]{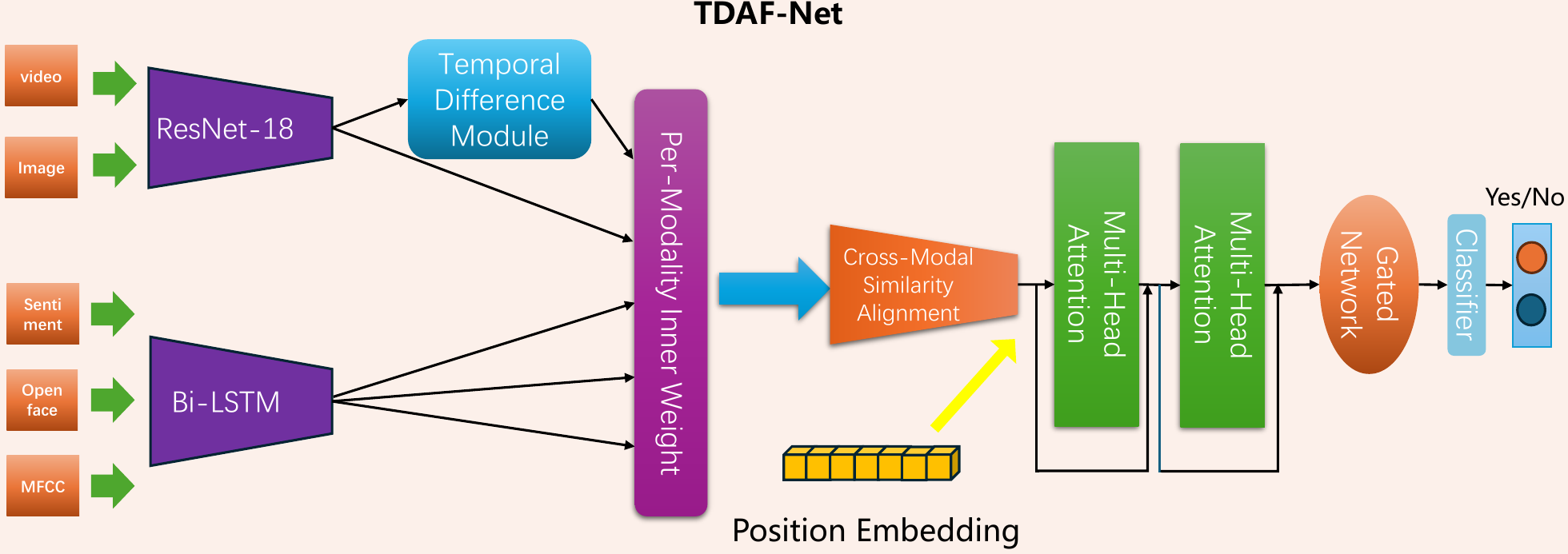}
	\caption{Network framework of Team ahrior.}
	\label{ahrior}
\end{figure}

TDAF-Net first integrates visual, audio, and textual cues to uncover subtle forgery traces that are invisible in single-modality analysis. Then, it deploys a Temporal Difference module to capture inter-frame discrepancies, such as inconsistent motions or abrupt scene changes. Meanwhile, it also deploys a multimodal Bi-LSTM to encode sentiment, facial expression, and MFCC-based audio features into sequential representations. Subsequently, a Per-Modality Inner Weight layer dynamically aggregates these heterogeneous features, emphasizing discriminative modalities, and a Cross-Modal Similarity Alignment module aligns visual and text-audio representations by learning their semantic correspondences, and generates fine-grained multimodal features that will be further refined.Finally, a Gated Network adaptively fuses the refined features and predicts the authenticity. The framework of the proposed method is shown in Fig. \ref{ahrior}.

\section{The First Domain Generalized Remote Physiological Measurement Challenge}

\subsection{Challenge Corpora}
The competition is conducted on five distinct datasets to rigorously evaluate physiological measurement and cross-domain generalization: UBFC-rPPG~\cite{bobbia2019unsupervised}, PURE~\cite{stricker2014non}, BUAA-MIHR~\cite{BUAA}, MMPD~\cite{MMPD}, and PhysDrive~\cite{wang2025physdrive}. Due to strict dataset permission restrictions, the official organizing committee does not provide the original files of the training and test datasets directly to participants. Instead, all participating teams must independently apply for and download the original files of the five datasets from their respective official websites. These datasets consist of facial videos capturing subtle skin color variations, accompanied by ground-truth physiological signals.

The UBFC-rPPG, PURE, and BUAA-MIHR datasets serve as the foundational corpora for baseline multi-domain feature learning. These datasets encompass diverse physiological recordings collected under various illumination conditions and subject states, capturing continuous remote photoplethysmography (rPPG) signals. To construct a unified large multi-domain training dataset for Phase 1, all samples from these three datasets are merged entirely. This aggregation is critical for enabling the models to learn domain-invariant features across varying camera sensors and baseline environmental setups.

The Multi-Domain Mobile Video Physiology Dataset (MMPD) is introduced to specifically assess and enhance the model's robustness against complex real-world artifacts. MMPD features a comprehensive collection of videos capturing varied skin tones, distinct lighting variations, and diverse motion artifacts. During the Phase 1 training process, exactly $50\%$ of the samples from the MMPD dataset are integrated with the UBFC-rPPG, PURE, and BUAA-MIHR datasets. Participants must complete model training based solely on this combined training set, and the use of any external data or additional training strategies is strictly prohibited.

We use the remaining $50\%$ of the samples from the MMPD dataset for Phase 1 evaluation. This phase specifically evaluates the model's generalization ability to unseen samples within the same domain after training on the multi-domain synthetic data. The specific division of training and test samples has been explicitly provided on Github to ensure standardized evaluation. After training, participants conduct performance evaluations on this designated test set.

For Phase 2 evaluation, we employ the PhysDrive dataset as the exclusive test set. This phase focuses on evaluating the generalization stability of the model to completely unseen domains under fixed weights. Participants may perform fine-tuning or re-training based on their Phase 1 models; however, after this process, the model weights must be strictly fixed. Based on these fixed weights, prediction tasks are completed on the specific PhysDrive test set. Submissions require an independent txt file for each test sample containing the temporal rPPG waveform data, seamlessly reflecting the model's capacity to overcome extreme domain shifts.

\subsection{Evaluation Metrics}
The performance of rPPG-based HR prediction models is evaluated by three core metrics, which are calculated on the predicted HR values ($\hat{y}_i$) and ground-truth HR values ($y_i$) derived from HR estimation via Fast Fourier Transform (FFT) method.

\paragraph{(1) Mean Absolute Error (MAE)}
MAE measures the average absolute deviation between the predicted HR and the ground-truth HR, reflecting the overall bias of the prediction model. The formula is defined as:
\begin{equation}
\text{MAE} = \frac{1}{N}\sum_{i=1}^{N} \left| \hat{y}_i - y_i \right|
\end{equation}
where $N$ denotes the number of HR pairs, $\hat{y}_i$ is the $i$-th predicted HR value, and $y_i$ is the $i$-th ground-truth HR value. A smaller MAE value indicates a more accurate HR prediction.

\paragraph{(2) Root Mean Square Error (RMSE)}
RMSE quantifies the square root of the average of the squared deviations between predicted and ground-truth HR, which penalizes large prediction errors more heavily than MAE. The formula is defined as:
\begin{equation}
\text{RMSE} = \sqrt{\frac{1}{N}\sum_{i=1}^{N} \left( \hat{y}_i - y_i \right)^2}
\end{equation}
A smaller RMSE value means the model has fewer large errors in HR prediction.

\paragraph{(3) Pearson's Correlation Coefficient ($r$)}
The Pearson's correlation coefficient $r$ measures the linear correlation (including strength and direction) between the predicted HR sequence and the ground-truth HR sequence. The formula is defined as:
\begin{equation}
r = \frac{\text{Cov}(\hat{y}, y)}{\sigma_{\hat{y}} \cdot \sigma_y} = \frac{\sum_{i=1}^{N} (\hat{y}_i - \bar{\hat{y}})(y_i - \bar{y})}{\sqrt{\sum_{i=1}^{N} (\hat{y}_i - \bar{\hat{y}})^2} \cdot \sqrt{\sum_{i=1}^{N} (y_i - \bar{y})^2}}
\end{equation}

where:
$\text{Cov}(\hat{y}, y)$ is the covariance between predicted HR sequence $\hat{y}$ and ground-truth HR sequence $y$;
$\sigma_{\hat{y}}$ is the standard deviation of $\hat{y}$;
$\sigma_y$ is the standard deviation of $y$;
$\bar{\hat{y}}$ is the mean value of $\hat{y}$;
$\bar{y}$ is the mean value of $y$.

The value of $r$ ranges from $-1$ to $1$. A value closer to $1$ indicates a strong positive linear correlation, a value closer to $-1$ indicates a strong negative linear correlation, and a value close to $0$ indicates no linear correlation. A higher absolute value of $r$ means the model can capture the dynamic variation of the real heart rate effectively.


\subsection{ Participation}

A total of 3 teams submitted their results by the end of the workshop competition \footnote{PhysDG 2026 Challenge Results: \url{https://www.codabench.org/competitions/12857/\#/results-tab}}. Results for Phase 1 and Phase 2 are reported in Tables \ref{tab:PhysDG_Phase1_results} and \ref{tab:PhysDG_Phase2_results}, respectively. We will introduce the approaches of some teams in the following part.

\begin{table}[h]
\centering
\caption{PhysDG Challenge Phase 1 Results.}
\begin{tabular*}{\linewidth}{@{\extracolsep{\fill}} c l c c c}
\toprule
\# & Team & RMSE & MAE & $r$ \\
\midrule
1 & GDMU\_ZZU & 8.06 & 3.20  & 0.86 \\
2 & RPM\_HFUT & 12.84 & 6.69  & 0.57 \\
3 & zin\_chou & 19.39 & 15.17 & 0.05 \\
\bottomrule
\end{tabular*}
\label{tab:PhysDG_Phase1_results}
\end{table}

\begin{table}[h]
\centering
\caption{PhysDG Challenge Phase 2 Results.}
\begin{tabular*}{\linewidth}{@{\extracolsep{\fill}} c l c c c}
\toprule
\# & Team & RMSE & MAE & $r$ \\
\midrule
1 & RPM\_HFUT & 15.06 & 10.61  & 0.26 \\
2 & zin\_chou & 24.05 & 17.68  & 0.06\\
3 & GDMU\_ZZU & 25.71 & 17.75  & 0.04 \\
\bottomrule
\end{tabular*}
\label{tab:PhysDG_Phase2_results}
\end{table}

\subsubsection{ Team RPM\_HFUT}
\begin{figure}[h]
	\centering
	\includegraphics[width=0.48\textwidth]{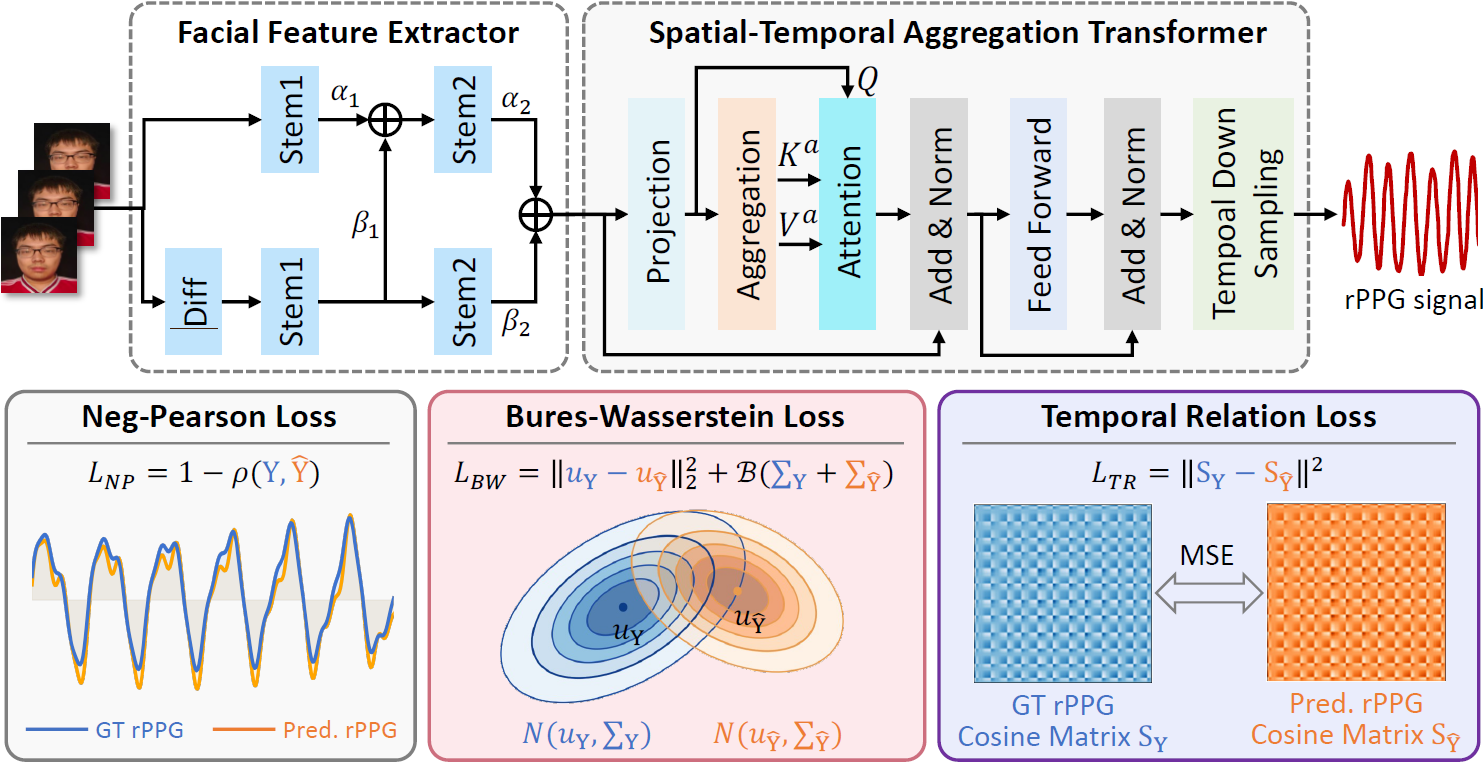}
	\caption{Network framework of Team RPM-HFUT.}
	\label{RPM-HFUT}
\end{figure}

The RPM\_HFUT team observes that existing deep rPPG methods typically rely on the negative Pearson correlation loss, which enforces only sample-level alignment and fails to explicitly capture the intrinsic statistical structure and temporal dependencies of physiological signals. This limitation becomes more evident under domain shifts, where variations in illumination, motion artifacts, and subject appearance may lead to predictions deviating from the true physiological distribution. To address this issue, they enhance the standard supervision with two complementary constraints. First, they introduce a Bures–Wasserstein loss to align the empirical mean and covariance of predicted and ground-truth rPPG signals, thereby enforcing distribution-level consistency. Second, they propose a temporal relation consistency loss that preserves pairwise dependencies across time steps via similarity matrix alignment, enabling the model to capture intrinsic rhythmic structures. The framework of the proposed method is shown in Fig. \ref{RPM-HFUT}.

\subsubsection{ Team GDMU\_ZZU}

\begin{figure}[h]
	\centering
	\includegraphics[width=0.48\textwidth]{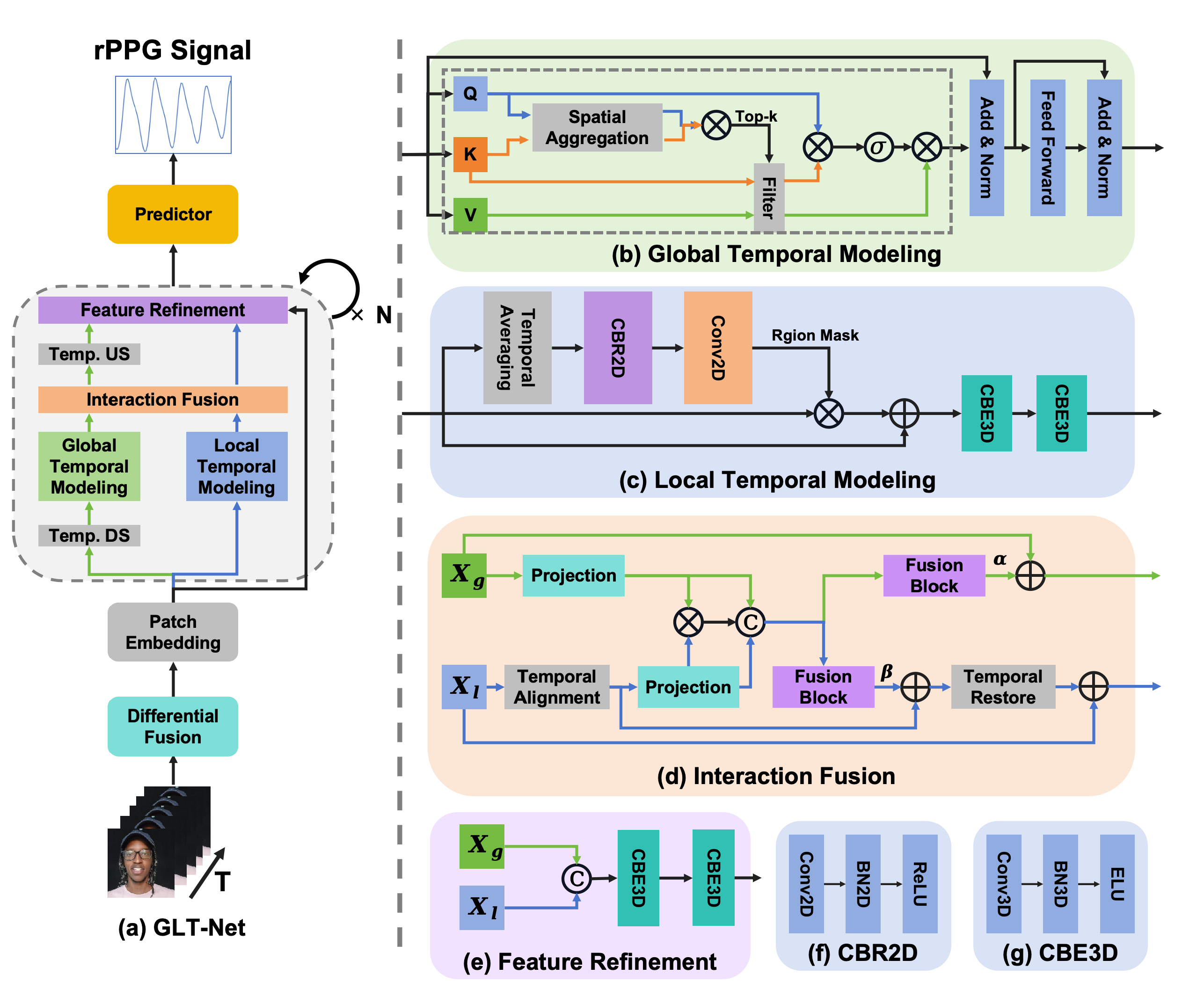}
	\caption{Network framework of Team GDMU\_ZZU.}
	\label{GDMU_ZZU}
\end{figure}

In rPPG estimation, pulse-related signals are typically weak and easily affected by background noise and non-physiological interference, while effective representation learning requires jointly modeling long-range periodic dependencies and short-term temporal dynamics. To address these challenges, Team GDMU\_ZZU proposes a dual-branch collaborative global-local temporal modeling network for rPPG estimation, as shown in Fig.\ref{GDMU_ZZU}. The model first employs Differential Fusion to integrate appearance information from raw frames with dynamic cues from inter-frame differences, then adopts a Transformer-based global temporal modeling branch to capture long-range periodic dependencies and stable pulse rhythms, together with a convolution-based local temporal modeling branch to strengthen short-term temporal dynamics; to promote effective collaboration between the two branches, an Interaction Fusion module is introduced to enable explicit cross-branch interaction and bidirectional residual refinement in a shared feature space. In addition, the local branch generates region masks from temporally averaged features to enhance features from physiologically informative regions while suppressing irrelevant interference, and finally, the fused spatiotemporal representation is fed into a temporal regression head to recover the target rPPG waveform.

\section{Conclusion}
In this paper, we present the SVC 2026 challenge, which focuses on modeling subtle visual signals through two tasks: multimodal deception detection and domain generalized rPPG estimation. By providing a unified evaluation framework, standardized datasets, and reproducible protocols, the challenge offers a systematic benchmark for studying robustness and cross-domain generalization. The results show that, despite recent progress, handling domain shifts and weak signal modeling remains difficult. We hope this challenge will encourage further research toward more stable and generalizable multimodal models for real-world applications.

{
    \small
    \bibliographystyle{ieeenat_fullname}
    \bibliography{main}
}


\end{document}